\documentclass[letterpaper, 10 pt, conference]{ieeeconf}  

\IEEEoverridecommandlockouts                              

\usepackage{tikz}
\usepackage{xcolor}

\usepackage{multirow}
\usepackage{tabularx}
\usepackage{tabularx,booktabs}
\newcolumntype{C}{>{\centering\arraybackslash}X} 
\usepackage{ctable}
\usepackage{adjustbox}

\usepackage{graphicx}
\usepackage{mdframed}
\usepackage{tcolorbox}

\usepackage[detect-all]{siunitx}
\usepackage{amsfonts} 
\usepackage{amsmath}

\usepackage{lipsum}

\usepackage{amssymb}
\usepackage{pifont}
\newcommand{\cmark}{\ding{51}}%
\newcommand{\xmark}{\ding{55}}%

\usepackage[english]{babel}
\usepackage{blindtext}

\title{\LARGE \bf
Detection of Condensed Vehicle Gas Exhaust in LiDAR Point Clouds 
}

\author{Aldi Piroli$^{1}$, Vinzenz Dallabetta$^{2}$, Marc Walessa$^{2}$, Daniel Meissner$^{2}$, \\Johannes Kopp$^{1}$, Klaus Dietmayer$^{1}$
\thanks{$^{1}$ Institute of Measurement, Control, and Microtechnology, Ulm University, Germany {\tt\small \{firstname.lastname\}@uni-ulm.de}}
\thanks{$^{2}$ BMW~AG, Petuelring 130, 80809~Munich,~Germany {\tt\small \{vinzenz.dallabetta, marc.walessa\}@bmw.de} and {\tt\small daniel.da.meissner@bmwgroup.com}}%
}

\newcommand\copyrighttext{%
	\footnotesize \copyright\,2022 IEEE. Personal use of this material is permitted. Permission from IEEE must be obtained for all other uses, in any current or future media, including reprinting/republishing this material for advertising or promotional purposes, creating new collective works, for resale or redistribution to servers or lists, or reuse of any copyrighted component of this work in other works.}%
\newcommand\copyrightnotice{%
	\begin{tikzpicture}[remember picture,overlay]%
	\node[anchor=south,yshift=10pt] at (current page.south) {\fbox{\parbox{\dimexpr\textwidth-2cm}{\copyrighttext}}};%
	\end{tikzpicture}%
	\vspace{-10pt}%
}

\begin{document}
\maketitle
\copyrightnotice

\thispagestyle{empty}
\pagestyle{empty}

\begin{abstract}
LiDAR sensors used in autonomous driving applications are negatively affected by adverse weather conditions.
One common, but understudied effect, is the condensation of vehicle gas exhaust in cold weather. 
This everyday phenomenon can severely impact the quality of LiDAR measurements, resulting in a less accurate environment perception by creating artifacts like ghost object detections.
In the literature, the semantic segmentation of adverse weather effects like rain and fog is achieved using learning-based approaches. 
However, such methods require large sets of labeled data, which can be extremely expensive and laborious to get. 
We address this problem by presenting a two-step approach for the detection of condensed vehicle gas exhaust.
First, we identify for each vehicle in a scene its emission area and detect gas exhaust if present. 
Then, isolated clouds are detected by modeling through time the regions of space where gas exhaust is likely to be present. 
We test our method on real urban data, showing that our approach can reliably detect gas exhaust in different scenarios, making it appealing for offline pre-labeling and online applications such as ghost object detection.
\end{abstract}

\section{INTRODUCTION}
\begin{figure}[t!]
    \centering
    \includegraphics[width=\columnwidth]{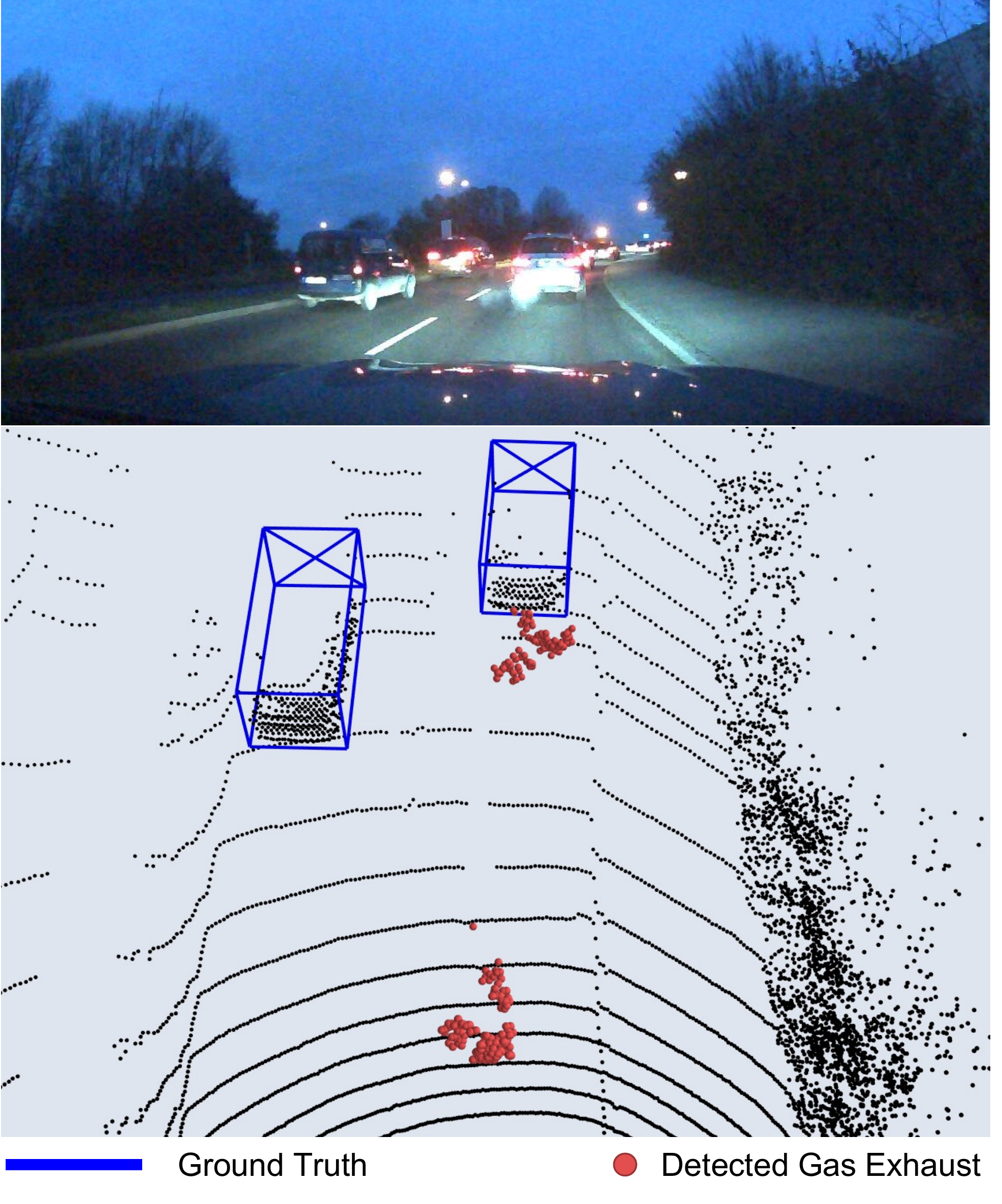}

    \caption{
    Detection results of our proposed method.
    In this example, two types of gas exhaust detections are present.
    The first, which we refer to as \textit{proximity gas exhaust}, is the gas cloud nearby the emitting vehicle. 
    The second, \textit{isolated gas exhaust}, is the cloud left behind by the same accelerating vehicle. }
    \label{Fig:teaser}
\end{figure}

LiDAR, together with radar and camera, are the principal sensors used in autonomous driving applications.
In recent years, a large body of literature has been proposed to process LiDAR point clouds. 
Different methods allow to semantically segment a point cloud ~\cite{milioto2019rangenet++, wu2018squeezeseg} or to detect and classify objects~\cite{shi2019pointrcnn, shi2020pv, lang2019pointpillars}. 
Although these methods have been proven to work reliably in good weather, only a few directly address performance in adverse weather conditions. 
This consideration is nonetheless critical since LiDAR sensors are negatively affected by adverse weather like snow, rain, and fog~\cite{jokela2019testing}.
In the context of autonomous driving, it is essential to reliably perceive and understand the surrounding environment independently of the weather conditions to guarantee safety and reliability.

This paper focuses on the understudied problem of condensed vehicle gas exhaust in LiDAR point clouds.
In cold weather, exhaust particles of combustion engines condense, forming a visible cloud before evaporating. 
Common Time-of-Flight (ToF) LiDAR sensors used in autonomous vehicles employ signals with a wavelength that can be reflected by these particles~\cite{de2009rf}. 
As shown in Fig.~\ref{Fig:teaser}, this leads to the introduction of potentially unwanted points in the returned measurement. 
Although they do not obstruct driving, such points can result in artifacts like ghost objects (Section~\ref{Sec:experiments}) that can negatively affect successive processing steps like sensor fusion and object tracking.
Considering that this phenomenon arises daily in scenarios like traffic lights, intersections or pedestrian crossings, gas exhaust poses an important challenge for the safety of upcoming autonomous vehicles and the environment they interact with. 

Although existing methods have shown promising results in dealing with adverse weather like fog, rain and snow, none address the problem of vehicle gas exhaust. 
Multi-sensor fusion methods like~\cite{bijelic2020seeing, pfeuffer2019robust} improve environment perception under adverse weathers, but the required additional sensors restrict their application.
For LiDAR-only methods, deep learning approaches have been used to detect rain, fog, smoke and dust in LiDAR point clouds~\cite{heinzler2020cnn, stanislas2021airborne}. 
However, they rely on large semantically annotated datasets, limiting the diversity of data during training.
In fact, the task of manual semantic labeling is exceptionally challenging for LiDAR point clouds, which can consist of more than \SI{100}{\kilo{}} points per scan. 
In adverse weather conditions, this task is made even harder by the irregularity of the weather effects, making the entire process time consuming and expensive.

We address the problems and limitations mentioned above by proposing the first method for detecting condensed vehicle gas exhaust in LiDAR point clouds.
In contrast to other approaches, our method works with common ToF LiDAR sensors and does not require labeled data.
We identify two cases in which gas exhaust is detected in LiDAR point clouds: proximity and isolated gas exhaust (Fig.~\ref{Fig:teaser}).
Proximity gas exhaust consists of clouds that remain in the vicinity of the emitting vehicle. 
Isolated gas exhaust clouds occur instead when a vehicle accelerates from a standstill position or  when a cloud drifts away from the emission position.
We address these two cases in separate steps. 
First, we detect proximity gas exhaust by determining the possible emission area for each vehicle present in the scene and identifying gas exhaust points if they are present.
Then, using past history measurements, we detect isolated gas exhaust by modeling through time the regions of space where gas exhaust clouds are likely to be present.
We extensively test our method in both the publicly available DENSE dataset~\cite{bijelic2020seeing} and a newly gathered one.
The results show that our approach can reliably detect gas exhaust in diverse urban scenarios, making our method appealing for both online gas exhaust detection and offline data labeling.
Furthermore, we show how our method can be used to improve the robustness of 3D object detections by detecting ghost objects caused by gas exhaust.  

Our main contributions can be summarized as follows:
\begin{itemize}
  \item We present a method for detecting vehicle gas exhaust in LiDAR point clouds, which does not require any labeled data or the use of additional sensors.
  \item We test our method on real data, showing that our approach can reliably identify vehicle gas exhaust in diverse urban scenarios.
  \item We show how our method can improve environment understanding by detecting ghost objects caused by gas exhaust clouds.
\end{itemize}

\begin{figure*}[t!]
    \centering
    \includegraphics[width=\textwidth]{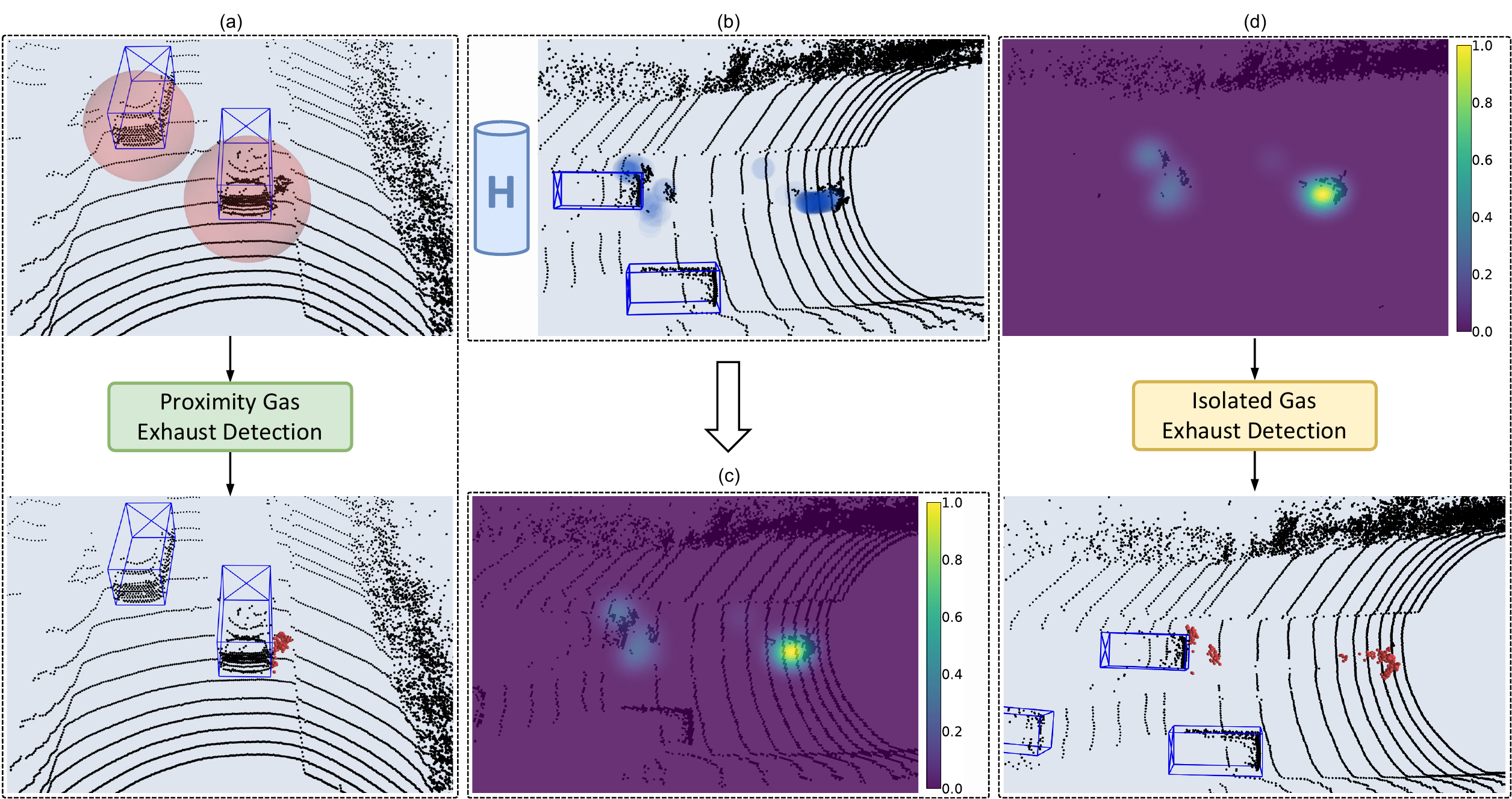}
    \caption{
    Overview of our proposed method.
    (a): for each vehicle bounding box in the scene, we identify proximity gas exhaust by defining the emission area $S$ (red spheres).
    We then cluster the detected gas exhaust points and save each detection $h$ (blue circles) in the detection history set $H$ (b).
    From $H$, we derive the 2D map $D$, where high values represent regions of space where gas exhaust is likely to be present. In (c), we show the map $D$ overlayed with the point cloud $P$.
    We use $D$ to detect the location of isolated gas exhaust points. 
    In (d), we show on the top the map $D$ overlayed with the point cloud $\widetilde{P}$ and on the bottom the detection results.
    We use blue lines for ground truth bounding boxes and red points for classified gas exhaust points.}
    \label{Fig:method}
\end{figure*}

\section{RELATED WORK}
\label{section:related_work}

\subsection{Point Cloud Processing}
In autonomous driving applications, it is important to distinguish between the drivable road and obstacles.
The authors of~\cite{himmelsbach2010fast} propose identifying road points by estimating local ground sectors and then inferring the global ground structure using 2D connected component labeling. 
In~\cite{paigwar2020gndnet}, ground points are detected using a deep neural network based on PointPillar~\cite{lang2019pointpillars} feature encoding.
Object detectors like~\cite{shi2019pointrcnn, shi2020pv, lang2019pointpillars} can be trained to predict different objects' positions, orientations, and classes in a point cloud.
An alternative approach consists on the semantic segmentation of each point in the point cloud. 
For example, networks like~\cite{milioto2019rangenet++, wu2018squeezeseg} project the point cloud in a range image and then use a convolutional neural network (CNN) to assign to each pixel a semantic class. 
Other methods like~\cite{qi2017pointnet++} avoid the quantization error of intermediate representations by directly processing the point cloud, resulting in slower inference times.

\subsection{Effects of Adverse Weather Conditions on LiDAR}
Adverse weather conditions like snow, rain, and fog can substantially degrade LiDAR measurements~\cite{jokela2019testing, heinzler2019weather}.
Snow and rain introduce random noise in the proximity of the sensor~\cite{charron2018noising, walz2021benchmark, filgueira2017quantifying}. 
At high speeds, the rain spray of leading vehicles can obstruct the field of view of the LiDAR sensor or cause ghost objects detections~\cite{walz2021benchmark}. 
In dense fog, the range of LiDAR sensors is significantly reduced by the scattering effect caused by the water particles that form it~\cite{bijelic2018benchmark}.
In cold weather, air requires less moisture to become saturated, hence heated gases exiting from a vehicle exhaust pipe can condense forming visible clouds~\cite{giechaskiel2019exhaust}. 
In~\cite{hasirlioglu2017effects}, the authors measure a vehicle emitting gas exhaust at different temperatures. 
The experiment shows that the returned number of points belonging to the vehicle decreases at low temperatures, while the number of points reflected by the condensed gas exhaust increases.
Moreover, multiple gas exhaust clouds can be detected in dynamic scenarios like a vehicle accelerating from a standstill position.

\subsection{Detection of Weather Effects in LiDAR Data}
Multi-sensor fusion approaches like~\cite{bijelic2020seeing,pfeuffer2019robust} integrate measurements from camera, radar and LiDAR sensors to improve the robustness of semantic segmentation and object detection algorithms in adverse weather conditions.
Denoising and classification algorithms based on spatial and reflectivity features have been used to detect rain and snow in point clouds~\cite{heinzler2019weather, charron2018noising, park2020fast}. 
However, gas exhaust is observed to return locally clustered points~\cite{hasirlioglu2017effects}, which makes pointwise filtering challenging to apply.
In~\cite{sallis2014air}, the authors propose classifying points belonging to smog, rain, and snow by the shape of the returned echo signals. 
However, this method does not apply to common LiDAR sensors, which only return the position and reflectivity of a measured point.
The work in~\cite{yoshida2017time} proposes an enhancement algorithm for ToF sensors to reduce the influence of gas exhaust in its gaseous state. 
The used sensor can produce both depth and infrared images, which are then used to determine the distribution of solid objects through the accumulation of multiple measurements. 
In~\cite{liang2020deep}, a method for detecting targets in adverse visibility conditions like fog and dust is proposed. The authors use a Frequency Modulated Continuous Wave (FMCW) LiDAR sensor which, in addition to range and reflectivity, can also measure the velocity of a point using the Doppler effect. 
In~\cite{heinzler2020cnn}, the authors propose a CNN approach for classifying rain and fog points in LiDAR point clouds.
To gather a large number of labeled data, they use a weather chamber where static scenarios are recreated using objects like mannequins, traffic cones and cars.
They then overlay artificially generated rain and fog and label the data using background subtraction.
Although this approach avoids the problem of manual labeling, it is severely limited by the required infrastructure. 
Moreover, the labeling strategy does not cover complex and dynamic urban settings, which are common in real scenarios.
A similar approach is presented in~\cite{stanislas2021airborne} to detect smoke and dust particles in robotic applications.


\section{METHOD}
In this section, we describe in detail our proposed method for detecting vehicle gas exhaust in LiDAR point clouds. 
In Fig.~\ref{Fig:method}, the overall structure of our approach is shown. 
\subsection{Proximity Gas Exhaust Detection}
\label{Sec:method_proximity_gas}
At time step $t$, we have a point cloud $P_t = \{p_0, ..., p_n \}$, where $p_i = (x,y,z,r)$ describes the position and reflectivity of a point. 
Additionally, we assume to have a set of 3D bounding boxes $B_t = \{b_1, ..., b_m\}$ that describe the position and class of the objects in the scene. 
This information can be acquired from 3D labels when performing offline gas exhaust detection or using a state-of-the-art object detector~\cite{shi2020pv, shi2019pointrcnn, lang2019pointpillars} during online detection.
We derive the point cloud $\widetilde{P_t} \subset P_t$, which contains all the points of $P_t$ excluding those in the bounding boxes $B_t$ and the ones belonging to the road surface.
Road points can be detected using methods like~\cite{himmelsbach2010fast, paigwar2020gndnet}.
For each bounding box $b_j$ classified as a vehicle, we compute the center back point $p_{\text{back}, j}$ of the box. 
Since most exhaust pipes are located on the rear end of vehicles, we define the possible exhaust area as the sphere $S$ centered in $p_{\text{back}, j}$ and radius $s$.
We assign a label $\ell$ to each point $p_i \in \widetilde{P}_t$ using the following rule:
\begin{equation}
\ell= \begin{cases}
\textit{Gas} &\text{$d(p_{\text{back}, j},p_i)  \leq s$,}\\
\textit{Other} &\text{otherwise,}
\end{cases}
\label{Eq:ball_query}
\end{equation}
where $d(p_{\text{back}, j},p_i)$ is the euclidean distance from a point $p_i$ and a vehicle back point $p_{\text{back}, j}$.
The result of this operation is the set $\widehat{G}_t = \{ p \in \widetilde{P}_t \mid \ell = Gas \}$.
To avoid falsely classifying points like protruding objects, nearby walls or road signs, we project the points in $\widehat{G}_t$ into a pillar grid~\cite{lang2019pointpillars} with cell size $d_x$ and $d_y$.
For each pillar $q_k$, we compute the average reflectivity $\widehat{r}_k$ of the points inside and the minimum distance from the ground $g_{k}$. 
Gas exhaust is composed of airborne particles which float above the ground and return low reflectivity values. 
In contrast, solid objects are usually connected to the ground and have high reflectivity (Fig.~\ref{Fig:stats}).  
We use these observations to derive the following label correction rule:
\begin{equation}
\label{Eq:label_corr}
q_k^{\ell}= \begin{cases}
\textit{Gas} & \text{if $\widehat{r}_k < t_r$ $\land$ $g_{k} > 0$,} \\
\textit{Other} &\text{otherwise,}
\end{cases}
\end{equation}
which assigns the label $\ell$ to the points inside $q_k$. Here, $t_r$ is a reflectivity threshold parameter.
Depending on the capability of the LiDAR sensor, other constraints like first and second echo return comparison can be used to strengthen the label correction rule.   
The final set of gas exhaust points in the proximity of emitting vehicles is  $G_t = \{ p \in \widetilde{P}_t \mid \ell = Gas \}$.

\subsection{Isolated Gas Exhaust Detection}
\label{Sec:method_isolated_gas}
Detecting  isolated gas exhaust clouds is a much more difficult challenge than proximity gas exhaust since we cannot assume that a vehicle is in the vicinity. 
We solve this problem by modeling through time the regions of space where gas exhaust is present. 

At time step $t$, in the case of  $G_t \neq \emptyset$,  we cluster the points in $G_t$ to identify different gas exhaust clouds. 
For each cluster, we compute the $(x,y)$ mean position $\mu$ and covariance matrix $\Sigma$. 
We then save this information in the detection history set $H = \{h_0, ..., h_k\}$ where $h_i = (t, \mu, \Sigma)$. 
We keep a detection $h \in H$ for $T$ time steps since gas exhaust evaporates after a certain time. 
Similar to~\cite{yin2021center}, we generate a 2D grid $D_t$ with cell sizes $d_x$ and  $d_y$, initialized with zero values. 
For each measurement  $h \in H$, we generate a 2D Gaussian distribution $\mathcal{N}(\mu, \Sigma)$ with $(\mu, \Sigma) \in h$. 
We truncate $\mathcal{N}$ using the three-sigma rule and project it into the grid $D_t$.
Then, for each cell of $D_t$ we accumulate the associated value of $\mathcal{N}$.
Finally, we normalize the values of $D_t$.
As shown in Fig.~\ref{Fig:method}, the result is a 2D map where high cell values are located in regions of space where gas exhaust is likely to be present.
We define this method based on the assumption that a set of points is likely to be gas exhaust if their position is near a previously detected gas exhaust cloud.
To classify isolated gas exhaust points, we assign the label $\ell = Gas$ to a point $p \in \widetilde{P}_t$ if the associated cell value in $D_t$ is greater than $0$.
As for proximity gas exhaust, we apply the label correction rule \eqref{Eq:label_corr} and derive the isolated gas exhaust set $\widetilde{G}_t = \{ p \in \widetilde{P}_t \mid \ell = Gas \}$.
The points in $\widetilde{G}_t$ are then clustered and also added to detection history set $H$. 
The final set of gas exhaust points in the point cloud $P_t$ is $G_t \cup \widetilde{G}_t$.


\section{EXPERIMENTS}
\label{Sec:experiments}

\subsection{Datasets and Evaluation Metrics}
\label{Sec:dataset_and_eval_metrics}
To evaluate our proposed method, we use the publicly available DENSE dataset~\cite{bijelic2020seeing}, which was recorded traveling from Germany to Sweden.
The dataset offers a large selection of urban environments under diverse weather conditions. 
Given that the dataset provides only 3D labels, we first select all of the scans where gas exhaust is visible  in the camera view ($113$ scans) and then semantically label them, assigning to each point the label \textit{Gas} or \textit{Other}.
The number of gas exhaust scans is limited because the dataset offers only a subset of scans for each recorded scene.
Tests on other publicly available datasets are also limited since most are recorded in good weather conditions, and only a few contain adverse weather scenes.
For this reason, we collect a new dataset containing a large number of gas exhaust points generated from preceding vehicles.
The dataset, which we will refer to as \textit{Urban Gas Exhaust}, was recorded in an urban environment while in cold weather.
The used sensor is a roof-top mounted LiDAR with $40$ vertical beams operating at $20$\si{\hertz}.
We semantically label the data ($1380$ scans) using the same convention as for DENSE, resulting in approximately \SI{96}{\kilo{}} \textit{Gas} labeled points. 

We evaluate our method as a binary classification problem, where positive predictions refer to points classified as \textit{Gas} and negative to \textit{Other}. 
As evaluation metrics, we use precision (P), recall (R), intersection-over-union (IoU) and mean-IoU (mIoU)~\cite{wu2018squeezeseg}.
A comparison to other methods is not possible since, to the best of our knowledge, no other approach is available to detect vehicle gas exhaust using a common ToF LiDAR sensor.
Furthermore, learning-based methods like~\cite{heinzler2020cnn} and~\cite{stanislas2021airborne} would require a large number of annotated data to be adapted to gas exhaust detection.

\begin{figure}[t!]
    \centering
    \includegraphics[width=\columnwidth]{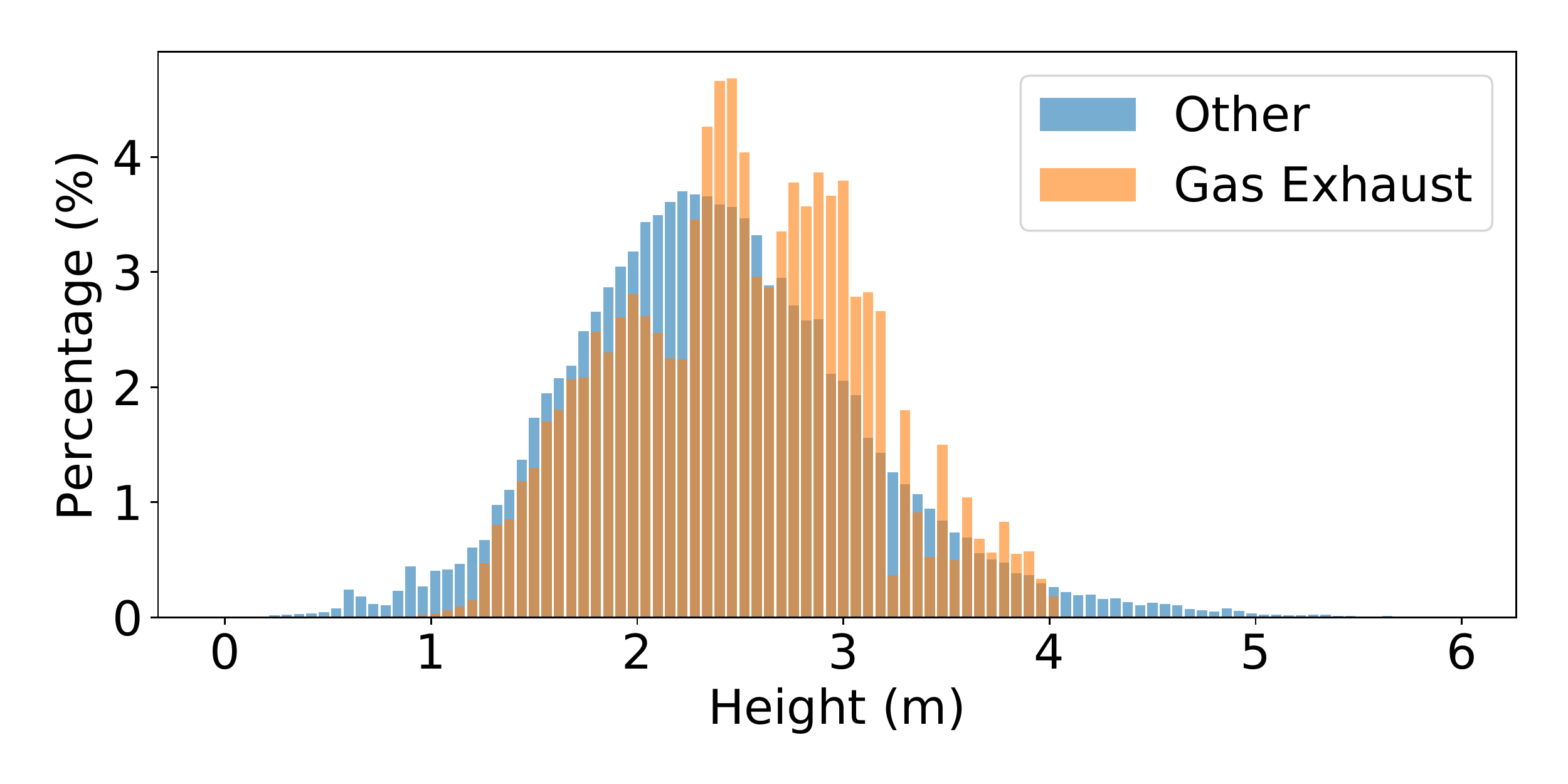}
    \includegraphics[width=\columnwidth]{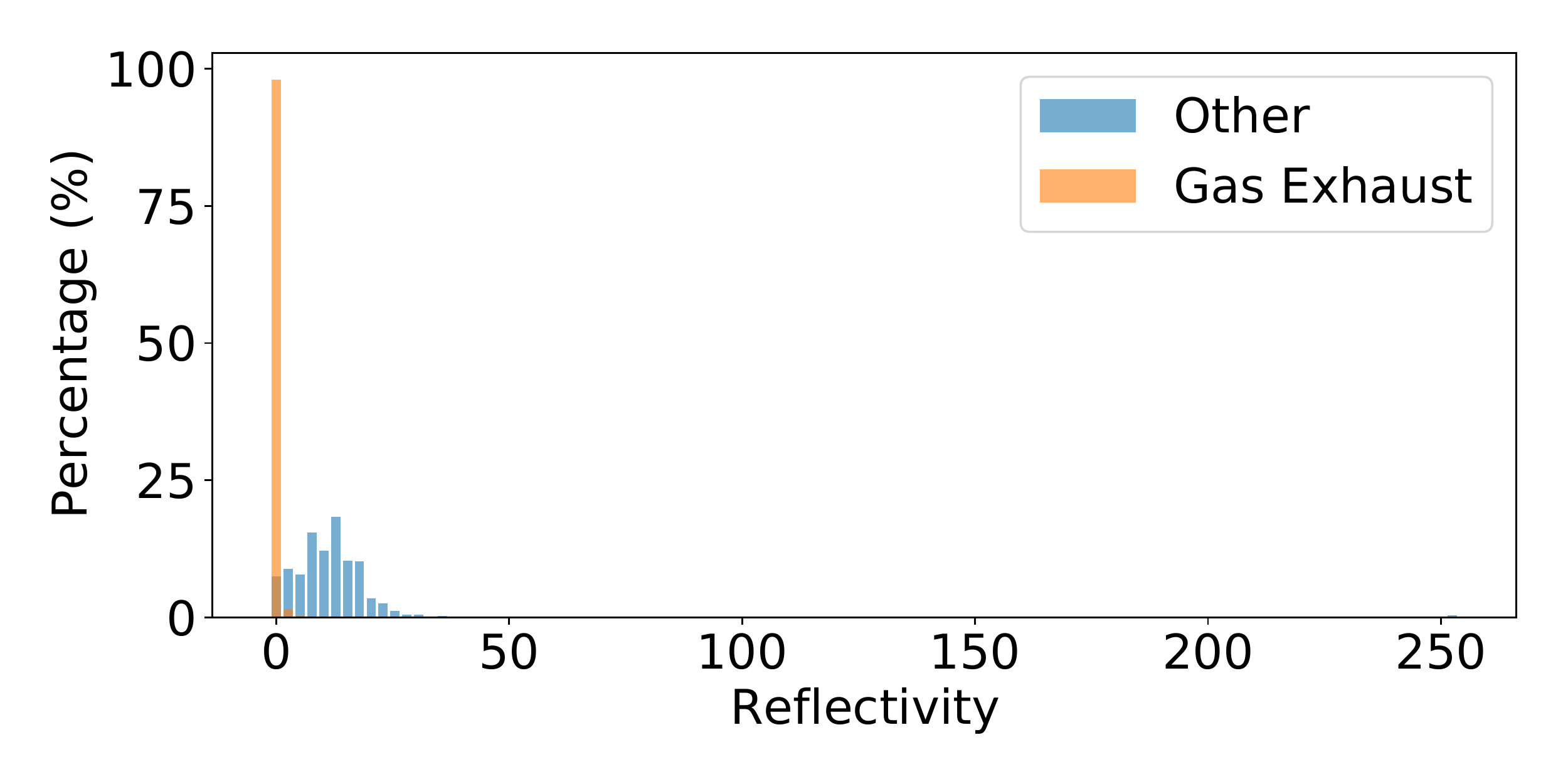}
    \caption{
    Height and reflectivity distribution of points in the Urban Gas Exhaust Dataset (Section~\ref{Sec:dataset_and_eval_metrics}). 
    The label ``Other'' refers to all the dataset points except for those belonging to road and gas exhaust.}
    \label{Fig:stats}
\end{figure}

\subsection{Implementation Details}
In our experiments, we use spheres $S$ with radii $s$ of 3m and 2m for the DENSE and Urban Gas Exhaust datasets, respectively.
We adopt a larger value of $s$ in DENSE since it does not offer continuous scans, and we can only rely on the proximity gas exhaust detection step.
In all experiments, we set the reflectivity threshold $t_r$ to $1\%$ of the maximum reflectivity value that the sensor can return since gas exhaust points have low reflectivity (Fig.~\ref{Fig:stats}).
From empirical observations, we set the parameter $T$, which represents the time a measurement is kept in $H$, to $150$ time steps.
For both the label correction rule \eqref{Eq:label_corr} and the derivation of the grid $D$, we use cell sizes $d_x$ and $d_y$ equal to \SI{0.1}{\meter}.

\subsection{DENSE Dataset}
\label{Sec:results_dense}
We start the evaluation by testing the proposed method on the DENSE dataset.
As mentioned in the previous sections, the dataset does not provide continuous scans limiting the use of algorithms that rely on sequential time frames like our isolated gas exhaust detection (Section~\ref{Sec:method_isolated_gas}).
Therefore, we test the performance of our proximity gas exhaust detection algorithm (Section~\ref{Sec:method_proximity_gas}), using as bounding boxes $B$ the provided dataset labels.
We make use of the large set of annotated data by also training two state-of-the-art 3D object detectors, PointRCNN~\cite{shi2019pointrcnn} and PVRCNN~\cite{shi2020pv}. We train them on the DENSE training set using the implementation provided by~\cite{openpcdet2020}.
Then, we generate predictions for each scan and use them as bounding boxes $B$.
To reduce the influence of errors in the object predictions, we increase the height, width and length of the boxes $B$ by $0.5$\si{\meter}.
We report the results in Table~\ref{Table:results_gas_eval_DENSE}.
When using ground truth labels as bounding boxes $B$ we reach a precision of $91.94\%$ and a recall of $37.39\%$, showing that our method can identify a large number of gas exhaust points in different urban scenarios while committing only few errors.
When using the detections of an object detector as bounding boxes $B$, we observe lower precision and recall values.
The lower precision can be associated with errors in the detector's predictions, whereas the lower recall with the increased bounding boxes sizes.
Nevertheless, we see that this approach can still benefit online applications with the best object detector achieving $63.06\%$ mIoU. 

\subsection{Urban Gas Exhaust Dataset}
We continue the evaluation by testing both steps of our method on the Urban Gas Exhaust dataset.
We use the 3D labels as bounding boxes $B$ needed to detect proximity gas exhaust (Section~\ref{Sec:method_proximity_gas}) and report the results in Table~\ref{Table:results_gas_eval_urban}.
Our method performs remarkably well, achieving precision and recall scores of $92.36\%$ and  $75.46\%$, respectively.
The obtained values allow for accurate detection of most gas exhaust points in the scene, improving the understanding of the autonomous vehicle surrounding environment.
These results are also significant when considering our method for an automatic pre-labeling of a dataset.
Our approach can be used to semantically label large datasets, drastically reducing the required manual effort.
Different from methods like~\cite{heinzler2020cnn}, our labeling strategy does not require large and expensive infrastructures.
Moreover, complex and dynamic scenarios can also be labeled, unlike in background subtraction labeling.
We highlight that the sensor used to record the Urban Gas Exhaust Dataset is different from the one used for DENSE. 
The experiments show that our method can be extended to different LiDAR sensors with only minimal adaptations. 

\begin{table}[t!]
\caption{Evaluation of the proposed method on the DENSE dataset.  The results refer to the semantic classification of vehicle gas exhaust points in LiDAR scans. 
Given the non-sequential nature of the provided data in DENSE, we perform detections using only the first step of our method (Section \ref{Sec:method_proximity_gas}).
The values are in percentage (\%).}
\label{Table:results_gas_eval_DENSE}
\centering
\begin{adjustbox}{max width=\columnwidth}
\begin{tabular}{@{}lccccc@{}}
\toprule
\multirow{2}{*}{Boxes ($B$)}& \multirow{2}{*}{Precision} & \multirow{2}{*}{Recall} & \multicolumn{2}{c}{IoU} & \multirow{2}{*}{mIoU} \\ \cmidrule(lr){4-5}
\multicolumn{1}{c}{}                                &                            &                         & \textit{Other}   & \textit{Gas}        &                       \\ \midrule
Labels                                              & 91.94                    & 37.39                 & 99.15    & 36.20    & 67.67               \\
PointRCNN \cite{shi2019pointrcnn}                                          & 73.56                    & 30.10                 & 98.95    & 27.16    & 63.06               \\
PVRCNN \cite{shi2020pv}                                                & 81.06                    & 25.50                 & 98.96    & 24.07    & 61.51               \\ \bottomrule
\end{tabular}%
\end{adjustbox}
\end{table}
\begin{table}[t!]
\caption{Evaluation of the proposed method on the Urban Gas Exhaust dataset. The results refer to the semantic classification of vehicle gas exhaust points in LiDAR scans. The values are in percentage (\%).}
\label{Table:results_gas_eval_urban}
\centering
\begin{adjustbox}{max width=\columnwidth}
\begin{tabular}{@{}ccccc@{}}
\toprule
\multirow{2}{*}{Precision}  & \multirow{2}{*}{Recall}     & \multicolumn{2}{c}{IoU}                                   & \multirow{2}{*}{mIoU}       \\ \cmidrule(lr){3-4}
                            &                             & \textit{Other}   & \textit{Gas}                         &                             \\ \midrule
\multicolumn{1}{r}{92.36} & \multicolumn{1}{r}{75.46} & \multicolumn{1}{r}{99.80} & \multicolumn{1}{r}{71.03} & \multicolumn{1}{r}{85.42} \\ \bottomrule
\end{tabular}%
\end{adjustbox}
\end{table}

\begin{figure*}[t!]
    \centering
    \includegraphics[width=\textwidth]{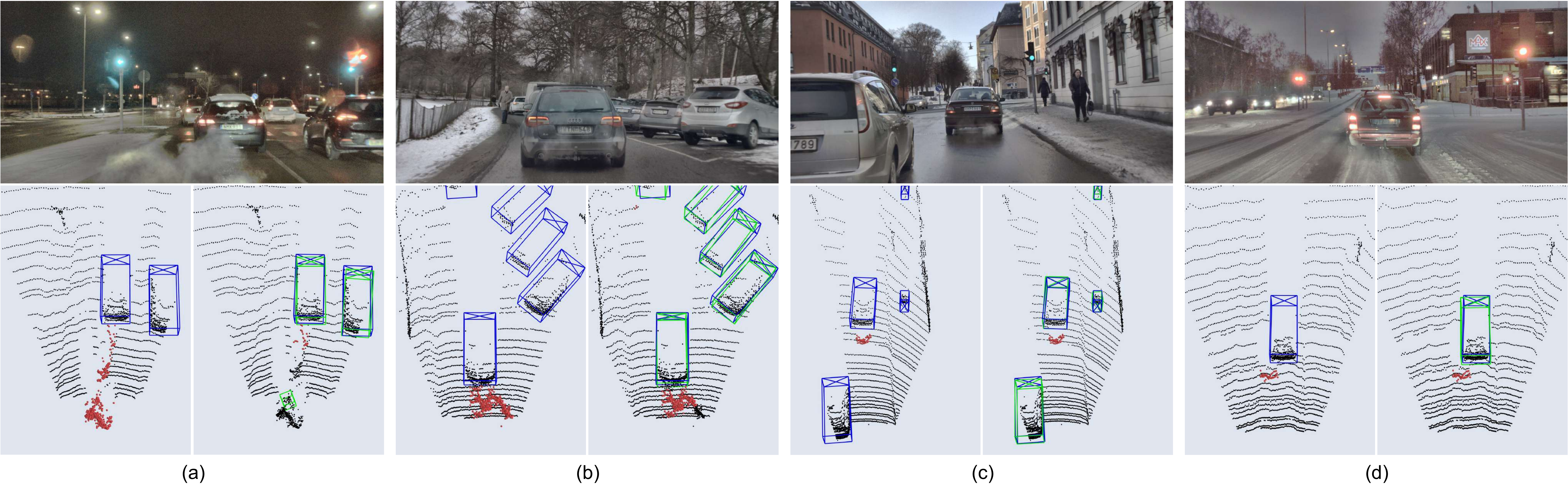}
    \caption{Qualitative results of the proximity gas exhaust detection step (Section~\ref{Sec:method_proximity_gas}) on the DENSE dataset.
    The blue lines represent the ground truth bounding boxes, the green the network predictions and the red points gas exhaust points. 
    For each scene, we show the camera view on top, the ground truth bounding boxes and gas exhaust semantic labels on the left, and the network predictions bounding boxes and gas exhaust point detections of our method on the right. 
    The predicted bounding boxes of Fig. (a) and (b) refer to the network PointRCNN, whereas Fig. (c) and (d) to PVRCNN.
    }
    \label{Fig:results_dense}%

    \end{figure*}

\subsection{Qualitative Evaluation}
In Fig.~\ref{Fig:results_dense} we show qualitative results of our proposed method on the DENSE dataset. 
In all scenarios, we use a state-of-the-art object detector to derive the bounding boxes $B$.
Although using only the first step of our method, a large number of the present gas exhaust points are detected.
In Fig.~\ref{Fig:results_dense}a, we see a failure case of our method. 
Here, a portion of the gas exhaust points remains undetected highlighting the importance of our second step, which directly addresses the detection of isolated gas exhaust clouds. 
In the examples, we see that gas exhaust is detected as dense clusters of points, which can result in artifacts like ghost object detections (Fig.~\ref{Fig:results_dense}a).
In Fig.~\ref{Fig:teaser},~\ref{Fig:method} and~\ref{Fig:ghost_obj}, we report examples of gas exhaust detections on the Urban Gas Exhaust dataset.
Here we can see that our method is successful in detecting both proximity and isolated gas exhaust clouds.

\subsection{Application to Ghost Object Detection}
\begin{figure}[t!]
    \centering
    \includegraphics[width=\columnwidth]{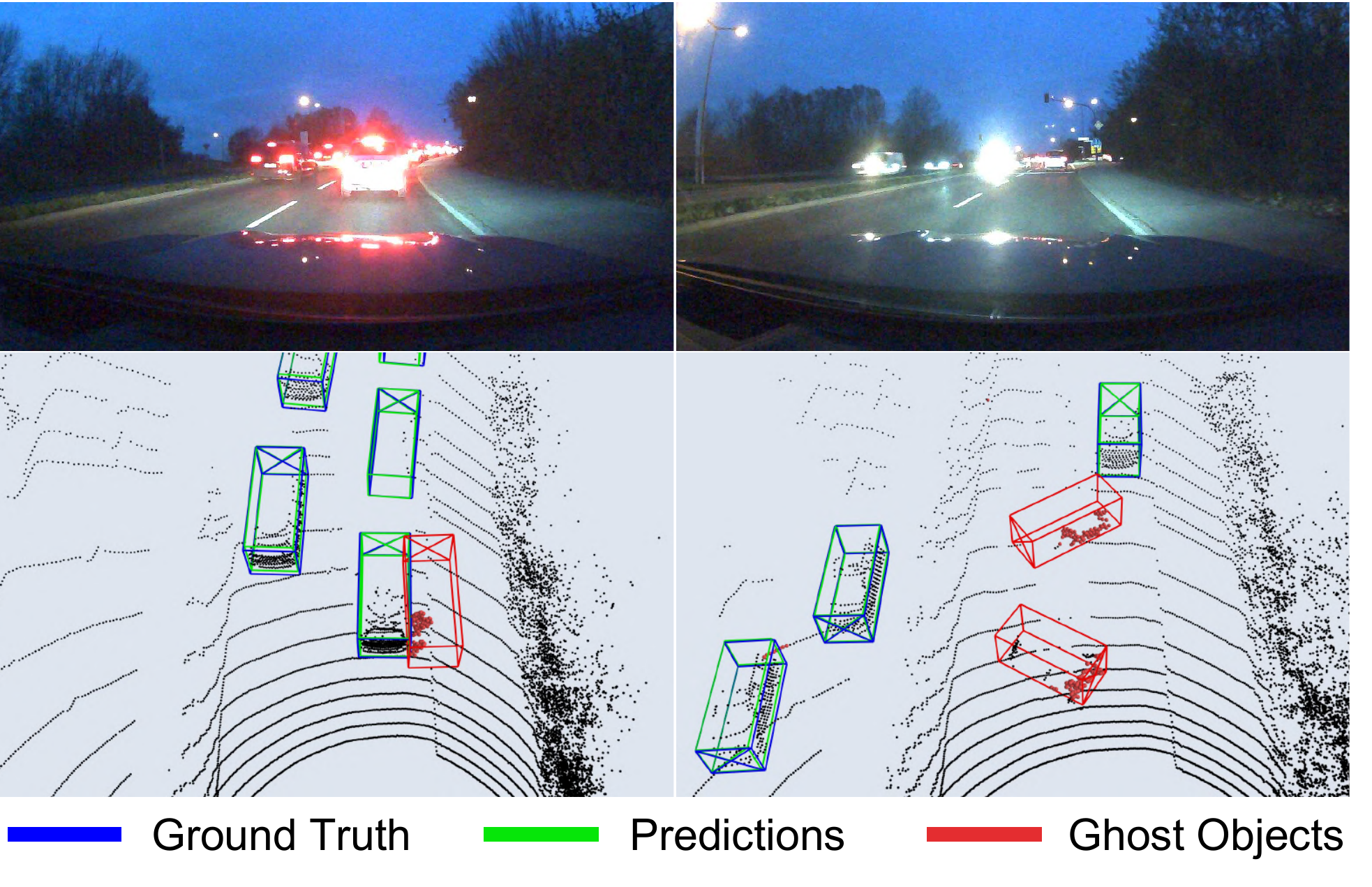}
    \caption{
    Ghost object detection performed using our proposed method.
    The predicted bounding boxes are generated using the object detector PointRCNN.}
    \label{Fig:ghost_obj}
\end{figure}

This section shows the application of our method to ghost object detections. 
We use the state-of-the-art object detector PointRCNN~\cite{shi2019pointrcnn} to perform detections on the Urban Gas Exhaust dataset. 
Due to the large size of annotated data that would be required to train the object detector, we use a pretrained model on the KITTI~\cite{geiger2013vision} dataset provided by~\cite{openpcdet2020}.
In the first step of our method, we select as bounding boxes $B$ the object detector predictions with confidence scores greater than $0.9$.
We then perform the detection of gas exhaust points in the point cloud using our proposed method. 
We classify a prediction as a ghost object if the majority of points inside the bounding box are classified as gas.
As we can see in Fig.~\ref{Fig:ghost_obj}, this simple strategy is successful in identifying ghost objects in the scene caused by gas exhaust. 

\section{ABLATION STUDIES}
\label{Sec:ablation}
In the following, we discuss the effect of our design choices and the effects on performance. 
All the experiments are performed on the Urban Gas Exhaust dataset using the dataset labels as 3D bounding boxes $B$.
\subsection{Gas Exhaust Area}
We start by first analyzing the effect on the performance of the parameter $s$, which expresses the radius of the sphere $S$ for the detection of proximity gas exhaust. 
As reported in Table~\ref{Table:ablation_radius}, with the increase of $s$ the recall increases whereas the precision decreases.
This result is expected since a larger radius allows to detect a higher number of proximity gas exhaust points. 
However, when passing from $2$\si{\meter} to $3$\si{\meter} the gain in recall only marginally increases by $0.22$ points, whereas the precision substantially decreases by $3.27$ points.

\subsection{Method's Features}
In Table~\ref{Table:ablation_method}, we report the results of activating or disabling features of the proposed method. 
First, we show the effect of disabling the label correction (Section~\ref{Sec:method_proximity_gas}), which decreases performance by $0.04\%$ mIoU points.
We can expect this result to differ when performing online detections using a state-of-the-art detector where the predicted object bounding boxes might not exactly match the ground truth.
When disabling the isolated gas exhaust detection stage (Section~\ref{Sec:method_isolated_gas}), we observe a decrease in performance of $15.32\%$ mIoU, showing that  a large portion of isolated gas exhaust points would remain undetected without this step.
Finally, we test the effect of saving isolated gas exhaust detections in the detection history set $H$.
When disabling this, we observe a decrease of $5.13\%$ mIoU from the best performance, which can be linked to instances where gas exhaust drifts away from the emission position.
By only considering proximity gas exhaust detections when deriving the grid $D$ (Section~\ref{Sec:method_isolated_gas}), portions of the space where gas exhaust might be present are not modeled.

\begin{table}[t!]
\caption{Ablation study of our proposed method performed using different sphere radius for the proximity gas exhaust detection (Section \ref{Sec:method_proximity_gas}). The values are in percentage (\%).}
\label{Table:ablation_radius}
\centering
\begin{adjustbox}{max width=\columnwidth}
\begin{tabular}{@{}cccccc@{}}
\toprule
\multirow{2}{*}{Radius ($s$)} & \multirow{2}{*}{Precision} & \multirow{2}{*}{Recall} & \multicolumn{2}{c}{IoU} & \multirow{2}{*}{mIoU} \\ \cmidrule(lr){4-5}
                        &                            &                         &\textit{Other}   & \textit{Gas}        &                       \\ \midrule
$1$\si{\meter}                           & \textbf{93.49}                   & 72.36                & 99.79   & 68.89   & 84.34              \\
$2$\si{\meter}                           & 92.36                   & 75.46                & \textbf{99.80}   & \textbf{71.03}   & \textbf{85.42}              \\
$3$\si{\meter}                           & 89.09                   & \textbf{75.68}               & 99.79   & 69.26   & 84.54              \\ \bottomrule
\end{tabular}%
\end{adjustbox}
\end{table}
\begin{table}[t!]
    \caption{Ablation study of our proposed method with components activated or deactivated. The values are in percentage (\%).}
    \label{Table:ablation_method}
    \centering
    \begin{adjustbox}{max width=\columnwidth}
        \begin{tabular}{@{}cccccc@{}}
            \toprule
            \multirow{2}{*}{\begin{tabular}[c]{@{}c@{}}Label \\ Correction\end{tabular}} & \multirow{2}{*}{\begin{tabular}[c]{@{}c@{}}Isolated \\ Gas Exhaust\end{tabular}} & \multirow{2}{*}{\begin{tabular}[c]{@{}c@{}}Second Stage \\ Memory Save\end{tabular}} & \multicolumn{2}{c}{IoU} & \multirow{2}{*}{mIoU}         \\ \cmidrule(lr){4-5}
                                                       &                                            &                                            & \textit{Other}   & \textit{Gas}                   &       \\ \midrule
            \xmark                                     & \cmark                                     & \cmark                                     & 99.80                   & 70.95                 & 85.38 \\
            \cmark                                     & \xmark                                     & \xmark                                     & 99.61                   & 40.59                 & 70.10 \\
            \cmark                                     & \cmark                                     & \xmark                                     & 99.74                   & 60.84                 & 80.29 \\
            \cmark                                     & \cmark                                     & \cmark                                     & \textbf{99.80}                   & \textbf{71.03}                 & \textbf{85.42} \\ \bottomrule
        \end{tabular}%
    \end{adjustbox}
\end{table}

\section{CONCLUSION AND FUTURE WORK}
In this paper, we presented a two-step method for condensed vehicle gas exhaust detection in LiDAR data.
Unlike other approaches for detecting adverse weather conditions, our method relies on common LiDAR sensors used in autonomous driving applications and does not depend on difficult to obtain labeled data. 
Results on real data show that our approach can successfully detect gas exhaust in different urban scenarios.
Our method can be used for pre-labeling applications, greatly reducing the manual effort of semantic labeling.
Moreover, its use can be extended to applications like ghost object detections, significantly improving the understanding of the surrounding environment in adverse weather conditions.  
In future work, we will explore the use of our approach for  other weather effects like rain spray.

\bibliographystyle{IEEEtran}
\bibliography{mybib}

\end{document}